\def\year{2022}\relax
\documentclass[letterpaper]{article} 
\usepackage{aaai22}  
\usepackage{times}  
\usepackage{helvet}  
\usepackage{courier}  
\usepackage[hyphens]{url}  
\usepackage{graphicx} 
\usepackage{natbib}  
\usepackage{caption} 
\DeclareCaptionStyle{ruled}{labelfont=normalfont,labelsep=colon,strut=off} 
\usepackage{algorithm}
\usepackage{algorithmic}
\usepackage[colorlinks=false]{hyperref}

\usepackage{newfloat}
\usepackage{listings}
\floatstyle{ruled}
\newfloat{listing}{tb}{lst}{}
\floatname{listing}{Listing}
\usepackage{array, makecell}
\usepackage{dblfloatfix}
\usepackage{arydshln}
\usepackage{cite}
\usepackage{dsfont}
\usepackage{bbm}
\usepackage{amsmath,amssymb,amsfonts}
\usepackage{graphicx}
\usepackage{array}
\usepackage{textcomp}
\usepackage{booktabs}
\usepackage[colorlinks=false]{hyperref}
\usepackage{xcolor}
\newtheorem{theorem}{Theorem}

\title{Hierarchical Optimal Transport for Unsupervised Domain Adaptation}

\author{
    Mourad El Hamri\textsuperscript{\rm 1,2,3},
    Younès Bennani\textsuperscript{\rm 1,2},
    Issam Falih\textsuperscript{\rm 2,4},
    Hamid Ahaggach\textsuperscript{\rm 2}
    \\
}
\affiliations{
    \textsuperscript{\rm 1}Laboratoire d'Informatique de Paris Nord, LIPN - CNRS UMR 7030, Université Sorbonne Paris Nord, France\\
    \textsuperscript{\rm 2}La Maison des Sciences Numériques - LaMSN, Université Sorbonne Paris Nord, France\\
    \textsuperscript{\rm 3}Dipartimento di Matematica, Universitá della Campania "Luigi Vanvitelli", Italy\\
    \textsuperscript{\rm 4}LIMOS - CNRS UMR 6518, Université Clermont Auvergne, France\\

    name.surname@\{sorbonne-paris-nord,uca\}.fr

}

\begin{document}

\maketitle

\begin{abstract}
In this paper, we propose a novel approach for unsupervised domain adaptation, that relates notions of optimal transport, learning probability measures and unsupervised learning. The proposed approach, HOT-DA, is based on a hierarchical formulation of optimal transport, that leverages beyond the geometrical information captured by the ground metric, richer structural information in the source and target domains. The additional information in the labeled source domain is formed instinctively by grouping samples into structures according to their class labels. While exploring hidden structures in the unlabeled target domain is reduced to the problem of learning probability measures through Wasserstein barycenter, which we prove to be equivalent to spectral clustering. Experiments on a toy dataset with controllable complexity and two challenging visual adaptation datasets show the superiority of the proposed approach over the state-of-the-art.
\end{abstract}
\section{Introduction}
\noindent Supervised learning is arguably the most widespread task of machine learning and has enjoyed much success on a broad spectrum of application domains \cite{kotsiantis2007supervised}. However, most supervised learning methods, are built on the crucial assumption that training and test data are drawn from the same probability distribution \cite{pan2009survey}. In real-world applications, this hypothesis is usually violated due to several application-dependent reasons: in computer vision, the presence or absence of backgrounds, the variation of acquisition devices or the change of lighting conditions introduce non-negligible discrepancies in data distributions \cite{saenko2010adapting}, in product reviews classification, the drift observed in the word distributions is caused by the difference of product category and the changes in word frequencies \cite{blitzer2007biographies}. These distributional shifts, will be likely to degrade significantly the generalization ability of supervised learning models. While manual labeling may appear as a feasible solution, such an approach is unreasonable in practice, since it is often prohibitively expensive to collect from scratch a new large high quality labeled dataset with the same distribution as the test data, due to lack of time, resources, or other factors,  and it would be an immense waste to totally reject the available knowledge on a different, yet related, labeled training set. Such a challenging situation has promoted the emergence of domain adaptation \cite{redko2019advances}, a sub-field of statistical learning theory \cite{vapnik2013nature}, that takes into account the distributional shift between training and test data, and in which the training set and test set distributions are respectively called source and target domains. There are two variants of domain adaptation problem, the unsupervised domain adaptation, where all the target data are unlabeled, and the semi-supervised domain adaptation, where few labeled target data are available. This paper deals with the challenging setting of unsupervised domain adaptation. 
\\
\\ Since the launching of domain adaptation theory, a large panoply of algorithms were proposed to deal with its unsupervised variant, and they can be roughly divided into shallow \cite{kouw2019review} and deep \cite{wilson2020survey} approaches. Most of shallow algorithms try to solve the unsupervised domain adaptation problem in two steps by first aligning the source and target domains to make them indiscernible, which then allows to apply traditional supervised methods on the transformed data. Such an alignment is typically accomplished through sample-based approaches which focus on correcting biases in the sampling procedure \cite{shimodaira2000improving,sugiyama2007direct} or feature-based approaches which focus on learning domain-invariant representations \cite{pan2010domain} and finding subspace mappings \cite{gong2012geodesic,fernando2013unsupervised}. Deep domain adaptation algorithms have also gained a renewed interest due to their feature extraction ability to learn more abstract and robust representations that are both semantically meaningful and domain invariant. \cite{ganin2016domain} is one of the most popular deep adaptative networks, which is based on the adversarial training procedure \cite{goodfellow2014generative} and directly derived from the seminal theoretical contribution in \cite{ben2007analysis}, its main idea is to embed domain adaptation into the representation learning process, so that the final classification decisions are made based on features that are both discriminative and invariant to domain changes.
\\
\\ More recent advances in domain adaptation are due to the theory of optimal transport \cite{villani2009optimal}, which allows to learn explicitly the least cost transformation of the source distribution into the target one. This idea was first investigated in the work of \cite{courty2016optimal}  where authors have successfully casted the domain adaptation problem into an optimal transport problem between shifted marginal distributions of the two domains, which then allows to learn a classifier on the transported data. Since then, several optimal transport based domain adaptation methods have emerged. In \cite{NIPS2017_0070d23b}, authors proposed to avoid the two-steps adaptation procedure, by aligning the joint distributions using a coupling accounting for the marginals and the class-conditional distributions shift jointly. Authors in \cite{redko2019optimal} performed  multi-source domain adaptation under the target shift assumption by learning simultaneously the class probabilities of the unlabeled target samples and the optimal transport plan allowing to align several probability distributions. The recent work of \cite{dhouib2020margin} derived an efficient optimal transport based adversarial approach from a bound on the target margin violation rate, to name a few.
\\
\\ A common denominator of these approaches is their ability to capture the underlying geometry of the data by relying on the cost function that reflects the metric of the input space. However, these optimal transport  based methods can benefit from not relying solely on such rudimentary geometrical information, since there is further important structural information that remains uncaptured directly from the ground metric, e.g., the local consistency structures induced by class labels in the source. The exploitation of these structures can induce some desired properties in domain adaptation like preserving compact classes during the transportation. It is, moreover, what led authors in \cite{courty2016optimal} to propose the inclusion of these structural information by adding a group-norm regularizer. Such additional structures, however, could not be induced directly by the standard formulation of optimal transport. To the best of our knowledge, \cite{alvarez2018structured} is the only work that has attempted to incorporate structural information directly into the optimal transport problem, without the need to add a regularized coefficient for expressing the class regularity constraints, by developing a nonlinear generalization of discrete optimal transport, based on submodular functions. However, the application of this method in domain adaptation only takes into account the available structures in the labeled source domain, by partitioning samples according to their class labels, while every target sample forms its own cluster. However, richer structures in the target domain can be easily captured differently, e.g., by grouping, and the incorporation of such  target structures directly in the optimal transport formulation can lead in our view to a significant improvement in the performance of domain adaptation algorithms.
\\
\\ In this paper, we address the existing limitations of the target-structure-agnostic algorithms mentioned above by proposing a principally new approach based on hierarchical optimal transport \cite{schmitzer2013hierarchical}. Hierarchical optimal transport, is an effective and efficient paradigm to induce structures in the transportation procedure. It has been recently used for different tasks such as multi-level clustering \cite{ho2017multilevel}, multimodal distribution alignment \cite{NEURIPS2019_e41990b1}, document representation \cite{NEURIPS2019_8b5040a8} and semi-supervised learning \cite{taherkhani2020transporting}. The relevance of this paradigm for domain adaptation is illustrated in Figure 1, where we show that the structure-agnostic Reg-OT and target-structure-agnostic OT-GL algorithms fail to always restrict the transportation of mass across instances of different structures, whereas, our Hierarchical Optimal Transport for Domain Adaptation (HOT-DA) model manages to do it correctly by leveraging the source and target domain structures simultaneously.
\begin{figure}[H]
	\centering
\includegraphics[width=0.48\textwidth]{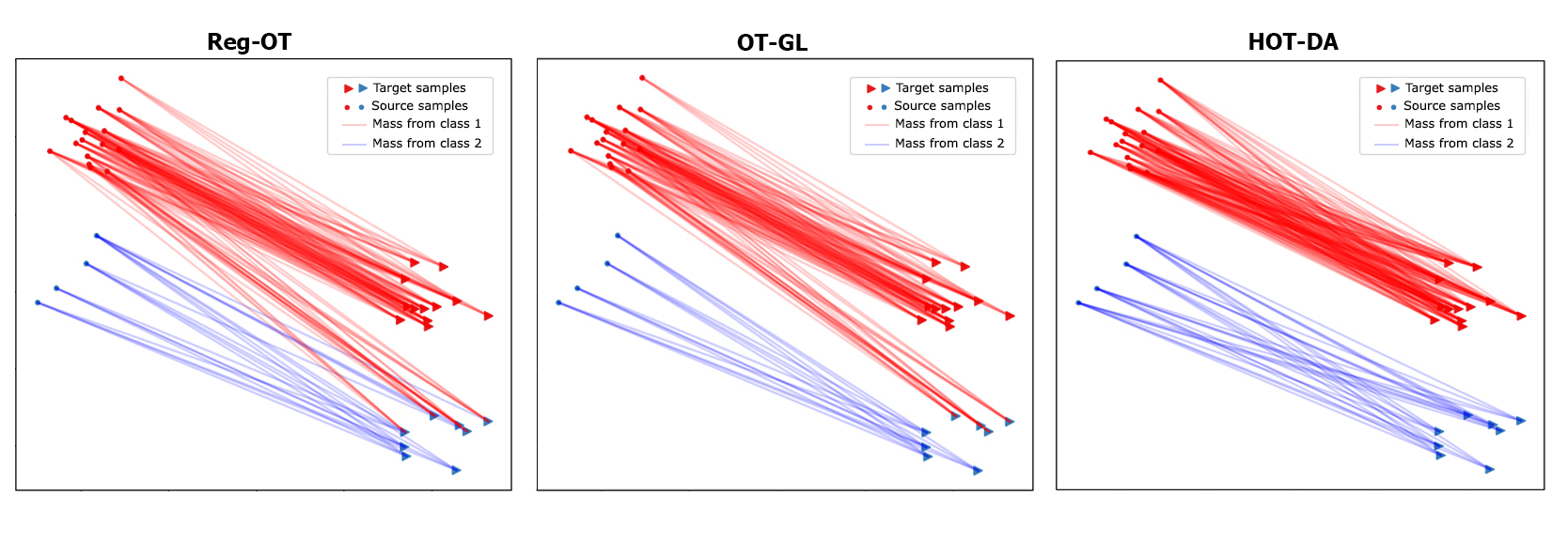}
	\caption{Illustration of the transportation obtained with structure-agnostic \cite{cuturi2013sinkhorn} and  target-structure-agnostic \cite{courty2016optimal} methods, and with HOT-DA.}
\end{figure}
\noindent To the best of our knowledge, the proposed approach is the first hierarchical optimal transport method for unsupervised domain adaptation and the first work to shed light on the connection between spectral clustering and Wasserstein barycenter. The rest of this paper is organized as follows: in the $2\textsuperscript{nd}$ section, we present a brief overview of unsupervised domain adaptation setup. In the $3\textsuperscript{rd}$ section, we detail the optimal transport problem and its hierarchical formulation,  then in the $4\textsuperscript{th}$ section, we elaborate the HOT-DA method. Finally, in the last section, we evaluate our algorithm on a toy dataset and two benchmark visual adaptation problems. 
\section{Unsupervised domain adaptation}
Let $\mathcal{X} = \mathbb{R}^d$ be an input space, $\mathcal{Y}=\{c_1,...,c_k\}$ a discrete label set consisting of $k$ classes, $\mathcal{S}$ and $\mathcal{T}$ two different probability distributions over $\mathcal{X} \times \mathcal{Y}$ called respectively the source and target domains. We have access to a set $S=\{(x_i,y_i)\}_{i=1}^n$ of $n$ labeled source samples drawn i.i.d. from the joint distribution $\mathcal{S}$ and a set $T=\{x_j\}_{j=1}^m$ of $m$ unlabeled target samples drawn i.i.d. from the marginal distribution $\mathcal{T}_\mathcal{X}$, of the joint distribution $\mathcal{T}$ over $\mathcal{X}$, more formally:
\begin{center}
    $S=\{(x_i,y_i)\}_{i=1}^n \sim (\mathcal{S})^n, \quad  T=\{x_j\}_{j=1}^m \sim (\mathcal{T}_\mathcal{X})^m$.
\end{center}
The aim of unsupervised domain adaptation algorithms is to infer a classifier $\eta: \mathcal{X} \to \mathcal{Y}$ with a low target risk:
\begin{center}
$\mathcal{R}_{\mathcal{T}}(\eta) = \underset{(x,y) \sim \mathcal{T}}{\mathbb{P}} (\eta(x) \neq y)$, 
\end{center}
under the distributional shift assumption $\mathcal{S} \neq \mathcal{T}$, while having no information about the labels $\{y_j\}_{j=1}^m$ of the target set $T$. In the rest, we design by the source domain interchangeably the distribution $\mathcal{S}$ and the labeled set $S$, and by the target domain, the distribution $\mathcal{T}$ and the unlabeled set $T$.
\section{Optimal Transport}
In this section we present the key concepts of optimal transport problem and its hierarchical formulation \cite{villani2009optimal}. 
\\ Optimal transport is a long-standing mathematical problem whose theory has matured over the time. Its roots can be traced back to the $18\textsuperscript{th}$ century, when the French mathematician Gaspard Monge introduced the following problem \cite{monge1781memoire}:  Let $(\mathcal{X},\mu)$ and $(\mathcal{Y},\nu)$ be two probability spaces, $c : \mathcal{X}\times\mathcal{Y} \to \mathbb{R}^+ $ a positive cost function over $\mathcal{X}\times\mathcal{Y}$, which represents the work needed to move a mass unit from $x \in \mathcal{X}$ to $y \in \mathcal{Y}$. The problem asks to find a measurable transport map  $\mathcal{T} : \mathcal{X} \to \mathcal{Y}$ that transports the mass represented by the probability measure $\mu$ to the mass represented by $\nu$, while minimizing the total cost of this transportation,
\begin{equation} (\mathcal{M}) \,\,\,\,\,\,\underset{\mathcal{T}}{\inf}\{\int_{\mathcal{X}}  c(x,\mathcal{T}(x)) d\mu(x) |  \mathcal{T}\#\mu = \nu \},   \end{equation}
where $\mathcal{T}\#\mu$ stands for the image measure of $\mu$ by $\mathcal{T}$. 
The problem of Monge $(\mathcal{M})$ is  quit  difficult,  since  it  is  not symmetric, and may not admit a solution, it is the case when $\mu$ is a Dirac measure and $\nu$ is not.
\\
\\ A long period of sleep followed  Monge’s formulation until the convex relaxation of the Soviet mathematician Leonid Kantorovitch in the thick of World War II \cite{kantorovich1942translocation}. This relaxed formulation, known as the Monge-Kantorovich problem $(\mathcal{MK})$ allows mass splitting and, in contrast to the formulation of Monge, it guarantees the existence of a solution under very general assumptions,
\begin{equation} (\mathcal{MK}) \,\,\,\,\,\,\underset{\gamma}{\inf} \{\, \int_{\mathcal{X}\times\mathcal{Y}} \, c(x,y) \, d\gamma(x,y) \,|\, \gamma \in \Pi(\mu,\nu)\, \}, \end{equation}
where $\Pi(\mu,\nu)$ is the transport plan set, constituted of joint probability measures $\gamma$ on $\mathcal{X}\times\mathcal{Y}$ with marginals $\mu$ and $\nu$, \\$\Pi(\mu,\nu) = \{ \gamma \in \mathcal{P}(\mathcal{X}\times\mathcal{Y})|proj_{\mathcal{X}}\#\gamma = \mu$, $proj_{\mathcal{Y}}\#\gamma = \nu$\}.
\\
\\ When $\mathcal{X}=\mathcal{Y}$ is a metric space endowed with a distance $d$,  a natural choice is to use it as a cost function, e.g., $c(x, y) = d(x, y)^p$ for $p \in {\left[1\,,+\infty\right[}$. Then, the problem $(\mathcal{MK})$ induces a metric between probability measures over $\mathcal{X}$, called the $p$-Wasserstein distance \cite{santambrogio2015optimal}, defined in the following way,  $\forall \mu,\nu \in \mathcal{P}(\mathcal{X})$:
\begin{equation}
    \mathrm{W}_{p}(\mu,\nu) = \underset{\gamma \in \Pi(\mu,\nu)}{\inf} (\int_{\mathcal{X}^{2}} d^{p}(x,y) \, d\gamma(x,y))^{1/p}, 
\end{equation}
In the discrete version of optimal transport, i.e., when the measures $\mu$ and $\nu$ are only available through discrete samples $X=(x_1,...,x_n)\subset\mathcal{X}  \, \text{and}  \, Y = (y_1,...,y_m) \subset \mathcal{Y}$, their empirical distributions can be expressed as
$\mu = \sum_{i=1}^n a_{i} \delta_{x_{i}}$ and $\nu = \sum_{j=1}^m b_{j} \delta_{y_{j}}$, where  $a=(a_1,...,a_n)$ and $b=(b_1,...,b_m)$ are vectors in the probability simplex $\sum_n$ and $\sum_m$ respectively. The cost function only needs to be specified for every pair $(x_i,y_j)_{\underset{1\leq j \leq m} {1\leq i \leq n}} \in X \times Y$ yielding a cost matrix $C \in  \mathcal{M}_{n \times m}(\mathbb{R}^{+})$. The optimal transport problem becomes then a linear program \cite{bertsimas1997introduction}, parametrized by the transportation polytope $U(a,b) = \{\gamma \in \mathcal{M}_{n \times m}(\mathbb{R}^{+}) \, | \, \gamma \mathds{1}_{m} = a \,\, \text{and} \,\, \gamma^{\mathbf{T}} \mathds{1}_{n} = b\}$, which acts as a feasible set, and the matrix $C$ which acts as a cost parameter. Thus, solving this linear program consists in finding a plan $\gamma^*$ that realizes:
\begin{equation}
(\mathcal{D}_\mathcal{MK}) \,\,\,\,\,\,\underset{\gamma \in U(a,b)}{\min}  \langle {\gamma},{C} \rangle _F,
\end{equation}
where $\langle.,.\rangle_F$ is the Frobenius inner product. In this case, the $p$-Wasserstein distance is defined as: $\mathrm{W}_{p}^p(\mu,\nu) = \langle {\gamma ^{*}},{C} \rangle _F$.
\\ Correlatively, a Wasserstein barycenter \cite{agueh2011barycenters} of $N$ measures $\{\nu_1,...,\nu_N\}$ in $\mathcal{P}(\mathcal{X})$ can be defined as a minimizer of the following functional $f$ over $\mathcal{P}(\mathcal{X})$:
\begin{equation}
f(\kappa) = \frac{1}{N} \sum_{i=1}^N \lambda_i \mathrm{W}_{p}^p(\kappa,\nu_i),
\end{equation}
where $\lambda_i$ are positive real numbers such that $\sum_{i=1}^N \lambda_i=1$.
\\
\\ As stated above, discrete optimal transport is a linear program, and thus can be solved exactly in $\mathcal{O}(r^3 log(r))$, where $r= \max(n,m)$, with the simplex algorithm or interior point methods \cite{pele2009fast}, which is a heavy computational price tag. Entropy-regularization \cite{cuturi2013sinkhorn} has emerged as a solution to the computational burden of optimal transport. The entropy-regularized discrete optimal transport problem is defined as follows:
\begin{equation}
(\mathcal{D}_\mathcal{MK}^{\,\varepsilon}) \,\,\,\,\,\,\underset{\gamma \in U(a,b)}{\min}  \langle {\gamma},{C} \rangle _F - \varepsilon \mathcal{H}(\gamma)   ,
\end{equation}
where  $\mathcal{H}(\gamma) = - \sum_{i=1}^n \sum_{j=1}^m \gamma_{ij} (\log(\gamma_{ij}) - 1)  $ is the entropy of $\gamma$. This regularization allows a faster computation of the optimal transport plan \cite{peyre2019computational} in $\mathcal{O}(r^2/\varepsilon^3)$ \cite{10.5555/3294771.3294958} via the iterative procedure of Sinkhorn algorithm \cite{knight2008sinkhorn}.
\\
\\ Hierarchical optimal transport is an attractive formulation that offers an efficient way to induce structural information directly into the transportation process \cite{schmitzer2013hierarchical}. Let $\mathcal{X}$ be a Polish metric space endowed with a distance $d$ and $\mathcal{P}(\mathcal{X})$ be the space of Borel probability measures on $\mathcal{X}$ equipped with the Wasserstein distance $\mathrm{W}_{p}$ according to (3). Since $\mathcal{X}$ is a Polish metric space, then $\mathcal{P}(\mathcal{X})$ is also a Polish metric space \cite{parthasarathy2005probability}. By a recursion of concepts, $\mathcal{P}(\mathcal{P}(\mathcal{X}))$ the space of Borel probability measures on $\mathcal{P}(\mathcal{X})$ is a Polish metric space, and will be equipped also with a Wasserstein metric $\mathrm{W}^{\,'}_{p}$ induced this time by the Wasserstein distance $\mathrm{W}_{p}$ which acts as the ground metric on $\mathcal{P}(\mathcal{X})$. More formally, let $\theta = (\mu_1,...,\mu_h)$ and  $\vartheta= (\nu_1,...,\nu_l)$ be two set of probability measures over $\mathcal{P}(\mathcal{X})$, i.e., $\theta, \vartheta \subset \mathcal{P}(\mathcal{X})$. The empirical distributions of $\theta$ and $\vartheta$ can be expressed respectively by $\phi,\varphi \in \mathcal{P}(\mathcal{P}(\mathcal{X}))$ as $\phi = \sum_{i=1}^h \alpha_{i} \delta_{\mu_{i}}$ and $\varphi = \sum_{j=1}^l \beta_{j} \delta_{\nu_{j}}$, where  $\alpha=(\alpha_1,...,\alpha_h)$ and $\beta=(\beta_1,...,\beta_l)$ are vectors in the probability simplex $\sum_h$ and $\sum_l$ respectively. Then, the hierarchical optimal transport problem between $\phi$ and $\varphi$ is:
\begin{equation} (\mathcal{HOT}) \,\,\,\,\,\,\underset{\Gamma \in U(\alpha,\beta)}{\min}  \langle {\Gamma},{\mathcal{W}} \rangle _F, 
\end{equation}
where the matrix $\mathcal{W} = (\mathrm{W}_{p}(\mu_i,\nu_j))_{\underset{1\leq j \leq l} {1\leq i \leq h}} \in  \mathcal{M}_{h \times l}(\mathbb{R}^{+})$, stands for the new cost parameter and $U(\alpha,\beta)$ represents the new transportation polytope defined in the following way: $U(\alpha,\beta) = \{\Gamma \in \mathcal{M}_{h \times l}(\mathbb{R}^{+}) \,\, | \,\, \Gamma \mathds{1}_{l} = \alpha \,\,\, \text{and} \,\,\, \Gamma^{\mathbf{T}} \mathds{1}_{h} = \, \beta\}$.
\section{HOT-DA: Hierarchical Optimal Transport for Unsupervised Domain Adaptation}
In this section, we introduce the proposed HOT-DA approach, that consists of three phases, the first one aims to learn hidden structures in the unlabeled target domain using Wasserstein barycenter, which we prove can be equivalent to spectral clustering, the second phase focuses on finding a one-to-one matching between structures of the two domains through the hierarchical optimal transport formulation, and the third phase involves transporting samples of each source structure to its corresponding target structure via the barycentric mapping.
\subsection{Learning unlabeled target structures through Wasserstein-Spectral clustering}
Samples in the source domain $S=\{(x_i,y_i)\}_{i=1}^n$ can be grouped into structures according to their class labels, but, data in the target domain $T=\{x_j\}_{j=1}^m$ are not labeled to allow us to identify directly such structures. Removing this obstacle cannot be accomplished without using some additional assumptions. In fact, to exploit efficiently the unlabeled data in the target domain, the most plausible assumption stems from the structural hypothesis based on clustering, where it is assumed that the data belonging to the same cluster are more likely to share the same label. This assumption constitutes the core nucleus for the first phase of our approach, which aims to prove that spectral clustering can be casted as a problem of learning probability measures with respect to Wasserstein barycenter. Our proof is based on three key ingredients: the equivalence between the search for a $2$-Wasserstein barycenter of the empirical distribution that represents unlabeled data and $k$-means clustering, the analogy between traditional $k$-means and kernel $k$-means and finally the connection between kernel $k$-means and spectral clustering. We derive from this result a novel algorithm able to learn efficiently hidden structures of arbitrary shapes in the unlabeled target domain.
\\
\\ Firstly, given $m$ unlabeled instances $\{x_1,...,x_m\} \subset \mathcal{X}$, $k$-means clustering \cite{macqueen1967some} aims to partition the $m$ samples into $k$ clusters $\Pi_k=\{\pi_1,...,\pi_k\}$ in which each sample belongs to the cluster with the nearest center. This results in a partitioning of the data space into Voronoi cells $(\text{Vor}_q)_{1\leq q \leq k}$ generated by the cluster centers $\tilde{C_k}=\{c_1,...,c_k\}$. The goal of $k$-means then is to minimize the mean squared error, and its objective is defined as:
\begin{equation}
    \underset{c_1,...,c_k}{\min} \frac{1}{m} \sum_{i=1}^m \lVert x_i - c_j \rVert^2,
\end{equation}
Let $\hat{\rho}_m = \sum_{i=1}^m \frac{1}{m} \delta_{x_{i}}$ be the empirical distribution of $\{x_1,...,x_m\}$. Since $\frac{1}{m} \sum_{i=1}^m \lVert x_i - c_j \rVert^2 = \mathbb{E}_{x \sim \hat{\rho}_m} \lVert x - \tilde{C_k} \rVert^2$, then according to \cite{NIPS2012_c54e7837}:
\begin{equation}
      \frac{1}{m} \sum_{i=1}^m \lVert x_i - c_j \rVert^2 = \mathrm{W}_{2}^2(\hat{\rho}_m,\pi_{\tilde{C_k}}\#\hat{\rho}_m),
\end{equation}
where $\pi_{\tilde{C_k}} : \mathcal{X} \to \tilde{C_k}$ is the projection function mapping each $x \in \text{Vor}_q \subset \mathcal{X}$ to $c_q$. Since $k$-means minimizes (9), it also finds the measure that is closest to $\hat{\rho}_m$ among those with support of size $k$ \cite{pollard1982quantization}. Which proves the equivalence between $k$-means and searching for a $2$-Wasserstein barycenter of $\hat{\rho}_m$ in $\mathcal{P}_k(\mathcal{X})$, i.e., a minimizer in $\mathcal{P}_k(\mathcal{X})$ of:
\begin{equation}
    f(\kappa) =  \mathrm{W}_{2}^2(\hat{\rho}_m,\kappa),
\end{equation}
Secondly, $k$-means suffers from a major drawback, namely that it cannot separate clusters that are nonlinearly separable in the input space. Kernel $k$-means \cite{scholkopf1998nonlinear} can
overcome this limitation by mapping the input data in $\mathcal{X}$ to a high-dimensional reproducing kernel Hilbert space $\mathcal{H}$ by a nonlinear mapping $\psi  : \mathcal{X} \to \mathcal{H}$, then the traditional $k$-means is applied on the high-dimensional mappings $\{\psi(x_1),...,\psi(x_m)\}$ to obtain a nonlinear partition. Thus, the objective function of kernel $k$-means can be expressed analogously to that of traditional $k$-means in (8):
\begin{equation}
    \underset{c_1,...,c_k}{\min} \frac{1}{m} \sum_{i=1}^m \lVert \psi(x_i) - c_j \rVert^2,
\end{equation}
Usually the nonlinear mapping $\psi(x_i)$ cannot be explicitly computed, instead, the inner product of any two mappings $\psi(x_i)^{T}\psi(x_j)$ can be computed by a kernel function $\mathcal{K}$. Hence, the whole data set in the high-dimensional space can be represented by a kernel matrix $K \in \mathcal{M}_{m}(\mathbb{R}^{+})$, where each entry is defined as: $K_{i,j} = \mathcal{K}(x_i,x_j) = \psi(x_i)^{T}\psi(x_j)$. 
\\
\\ Thirdly, according to \cite{zha2001spectral}, the objective function of kernel $k$-means in (11) can be transformed to the following spectral relaxed maximization problem:
\begin{equation}
    \underset{Y^{T}Y=I_{k}, Y \geq 0}{\max} trace(Y^{T}KY),
\end{equation}
On the other hand, spectral clustering has emerged as a robust approach for data clustering \cite{shi2000normalized,ng2002spectral}. Here we focus on the normalized cut for $k$-way clustering objective function  \cite{gu2001spectral,stella2003multiclass}. Let $G = (V, E, \tilde{K})$ be a weighted graph, where $V=\{x_1,...,x_m\}$ is the vertex set, $E$ the edge set, and $\tilde{K}$ the affinity matrix defined by a kernel $\tilde{\mathcal{K}}$. The $k$-way normalized cut spectral clustering aims to find a disjoint partition $\{V_1,...,V_k\}$ of the vertex set $V$, such that:
\begin{equation}
    \underset{V_1,...,V_k}{\min} \sum_{l=1}^k linkratio(V_l, \overline{V_l}),
\end{equation}
where $linkratio(V_l, \overline{V_l})=\frac{links(V_l,\overline{V_l})}{degree(V_l)}=\frac{\sum_{i \in V_l}\sum_{j \in \overline{V_l}}\tilde{K}_{ij}}{\sum_{i \in V_l}\sum_{j \in V}\tilde{K}_{ij}}$.
\\
\\ Following \cite{dhillon2004kernel,ding2005equivalence}, the minimization in (13) can be casted as:
\begin{equation}
    \underset{Z^{T}Z=I_{k}, Z \geq 0}{\max} trace(Z^{T}\tilde{D}^{-1/2}\tilde{K}\tilde{D}^{-1/2}Z),
\end{equation}
where $\tilde{D}$ is the degree matrix of the graph $G$. Thus, the maximization problem in (14) is identical to the spectral relaxed maximization of kernel $k$-means clustering in (12) when equipped with the kernel matrix $K=\tilde{D}^{-1/2}\tilde{K}\tilde{D}^{-1/2}$.
\\
\\ According to the three-dimensional analysis above, we can now give the main result in the $1\textsuperscript{st}$ phase of our method:
\begin{theorem}
Spectral clustering using an affinity matrix $\tilde{K}$ is equivalent to the search for a  $2$-Wasserstein barycenter of $\hat{\varrho}_m = \sum_{i=1}^m \frac{1}{m} \delta_{\xi (x_{i})}$ in the space of probability measures with support of size $k$, where $\xi$ is a nonlinear mapping corresponding to the kernel matrix $K=\tilde{D}^{-1/2}\tilde{K}\tilde{D}^{-1/2}$.
\end{theorem}
In the sequel, we will refer to the search for a $2$-Wasserstein barycenter of $\hat{\varrho}_m$ as Wasserstein-Spectral clustering, and we will use it to learn $k$ hidden structures in the unlabeled target domain $T$. There are fast and efficient algorithms to perform Wasserstein-Spectral clustering \cite{cuturi2014fast}, which offers an alternative to the popular spectral clustering algorithm of \cite{ng2002spectral}.
\\
\\ The theoretical result in \text{Theorem 1} is confirmed by experiments, this is illustrated in Figure 2, where we show that Wasserstein-Spectral clustering performs identically to the traditional spectral clustering, and that both are effective at separating clusters that are nonlinearly separable, whereas $k$-means fails to separate data with non-globular structures.
\begin{figure}[H]
	\centering
\includegraphics[width=0.48\textwidth]{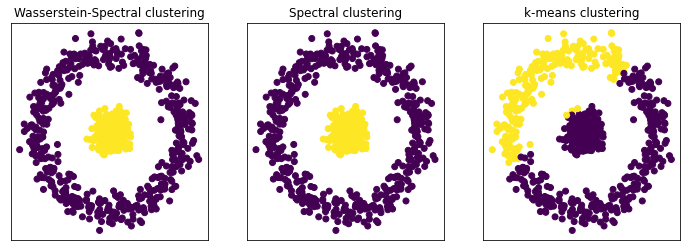}
	\caption{Comparison of Wasserstein-Spectral clustering, spectral clustering and $k$-means on Two-Circles dataset.}
\end{figure}
\subsection{Matching source and target structures through hierarchical optimal transport}
Optimal transport offers a well-founded geometric way for comparing probability measures in a Lagrangian framework, and for inferring a matching between them  as an inherent part of its computation. Its hierarchical formulation has inherited all these properties with the extra benefit of inducing structural information directly without the need to add any regularized term for this purpose, as well as the capability to split a sophisticated optimization surface into simpler ones that are less subject to local minima, and the ability to benefit from the entropy-regularization. Hence the key insight behind its use in the second phase of our method.
\\
\\ To use an appropriate formulation for hierarchical optimal transport, samples in the source domain $S=\{(x_i,y_i)\}_{i=1}^n$ must be partitioning according to their class labels  $y_i \in  \mathcal{Y}=\{c_1,...,c_k\}$ into $k$ classes $\{C_1,...,C_k\}$. The empirical distributions of these structures can be expressed using discrete measures $\{\mu_1,...,\mu_k\} \subset \mathcal{P}(\mathcal{X})$ as follows:
\begin{equation}
    \mu_h = \sum_{i=1/ x_i \in C_h}^n a_i \delta_{x_i}, \quad \forall h \in \{1,...,k\}
\end{equation}
Similarly, samples in the target domain $T=\{x_j\}_{j=1}^m$ are grouped in $k$ clusters $\{Cl_1,...,Cl_k\}$ using Wasserstein-Spectral clustering in the first phase. The empirical distributions of these structures can be expressed using discrete measures $\{\nu_1,...,\nu_k\} \subset \mathcal{P}(\mathcal{X})$ in the following way:
\begin{equation}
    \nu_l = \sum_{j=1/ x_j \in Cl_l}^m b_j \delta_{x_j}, \quad \forall l \in \{1,...,k\}
\end{equation}
Under the assumption that $S$ and $T$ are two collections of independent and identically distributed samples, the weights of all instances in each structure are naturally set to be equal:
\begin{center}
    $a_i = \frac{1}{\lvert C_h \rvert} \quad \text{and}  \quad b_j = \frac{1}{\lvert Cl_l \rvert}, \quad \forall h,l \in \{1,...,k\}$
\end{center}
All the labeled data in the source domain $S$ and the unlabeled data in the target domain $T$ can be seen in a hierarchical paradigm as a collection of classes and clusters. Thus, the distribution of $S$ and $T$ can be expressed respectively as a measure of measures $\phi$ and $\varphi$ in $\mathcal{P}(\mathcal{P}(\mathcal{X}))$ as follows:
\begin{equation}
    \phi = \sum_{h=1}^k \alpha_{h} \delta_{\mu_{h}} \quad \text{and} \quad \varphi = \sum_{l=1}^k \beta_{l} \delta_{\nu_{l}},
\end{equation}
where $\alpha=(\alpha_1,...,\alpha_k)$ and $\beta=(\beta_1,...,\beta_k)$ are vectors in the probability simplex $\sum_k$. The weights $\alpha_h$ and $\beta_l$ reflect the cardinality of the class $C_h$ in $S$ and the  cluster $Cl_l$ in $T$:
\begin{center}
    $\alpha_h = \frac{\lvert C_h \rvert}{n} \quad \text{and}  \quad \beta_l = \frac{\lvert Cl_l \rvert}{m}, \quad \forall h,l \in \{1,...,k\}$
\end{center}
To learn the correspondence between classes and clusters, we formulate an entropy-regularized hierarchical optimal transport problem between $\phi$ and $\varphi$ in the following way:
\begin{equation} (\mathcal{HOT\textbf{-}DA}) \,\,\,\,\,\,\underset{\Gamma \in U(\alpha,\beta)}{\min}  \langle {\Gamma},{\mathcal{W}} \rangle _F - \varepsilon \mathcal{H}(\Gamma), 
\end{equation}
where $U(\alpha,\beta)$ represents the transportation polytope: $U(\alpha,\beta) = \{\Gamma \in \mathcal{M}_{k}(\mathbb{R}^{+}) \,\, | \,\, \Gamma \mathds{1}_{k} = \alpha \,\,\, \text{and} \,\,\, \Gamma^{\mathbf{T}} \mathds{1}_{k} = \beta\}$.
$\mathcal{W} = (\mathcal{W}_{h,l})_{{1\leq h,l \leq k}} \in  \mathcal{M}_{k}(\mathbb{R}^{+})$ stands for the cost matrix, whose each matrix-entry $\mathcal{W}_{h,l}$ is defined as the $p$-Wasserstein distance between the measures $\mu_h$ and $\nu_l$, with:
\begin{equation}
    \mathcal{W}_{h,l}^p = \mathrm{W}_{p}^p(\mu_h,\nu_l) = \langle {\gamma ^{*,\varepsilon_{'}}_{h,l}},{\mathcal{C}_{h,l}} \rangle _F, 
\end{equation}
where $\mathcal{C}_{h,l}$ is the cost matrix of pairwise squared-Euclidean distances between elements of $C_h$ and $Cl_l$, and $\gamma ^{*,\varepsilon_{'}}_{h,l}$ is the regularized optimal transport plan between $\mu_h$ and $\nu_l$.
\\
\\ The optimal transport plan $\Gamma^*_\varepsilon$ of (18) can be interpreted as a soft multivalued matching between $\phi$ and $\varphi$ as it provides the degree of association between classes $\{C_1,...,C_k\}$ in the source domain $S$ and clusters $\{Cl_1,...,Cl_k\}$ in the target domain $T$. Then, the symmetric one-to-one matching relationship $(\widehat{=})$ between each class $C_h$ and its corresponding cluster $Cl_l$ can be inferred by hard assignment from $\Gamma^*_\varepsilon$:
\begin{equation}
    C_h  \widehat{=} \, Cl_l \mid l = \underset{j=1,...,k}{argmax} \, \Gamma^*_\varepsilon(h,j), \quad \forall h \in \{1,...,k\}
\end{equation}
\subsection{Transporting source to target structures through the barycentric mapping}
Besides being a means of comparison and matching, optimal transport has the asset of performing, as an intrinsic property of its transportation quiddity, an alignment of probability measures. Hence the main underlying idea of this phase.
\\ Once the correspondence between source and target structures has been determined according to the one-to-one matching relationship $(\widehat{=})$ in (20), the source samples in each class $C_h$ have to be transported to the target samples in the corresponding cluster $Cl_l$. This transportation can be handily expressed for each instance $x_i$ in $C_h$ with respect to the instances in $Cl_l$ as the following barycentric mapping \cite{reich2013nonparametric,ferradans2014regularized,courty2016optimal}:
\begin{equation}
   \widetilde{x_i} =  \underset{x \in \mathcal{X} }{argmin}  \sum_{j=1/ x_j \in Cl_l}^m \gamma^{*,\varepsilon_{'}}_{h,l}(i,j) \lVert x - x_j \rVert^2 
\end{equation}
where $\widetilde{x_i}$ is the image of $x_i$ in the region occupied by $Cl_l$ on the target domain, and $\gamma^{*,\varepsilon_{'}}_{h,l}$ is the optimal transport plan between $\mu_h$ and $\nu_l$ already computed in (19). The barycentric mapping can be formulated for each class $C_h$ as follows:
\begin{equation}
    \widetilde{C_h} = diag(\gamma^{*,\varepsilon_{'}}_{h,l}\mathds{1}_{\lvert Cl_l \rvert})^{-1}\gamma^{*,\varepsilon_{'}}_{h,l}Cl_l \quad \forall h \in \{1,...,k\}
\end{equation}
While samples in $C_h$ and $Cl_l$ are drawn i.i.d. from $\mu_h$ and $\nu_l$, then this mapping can be casted as a linear expression:
\begin{equation}
    \widetilde{C_h} =  \lvert C_h \rvert \gamma^{*,\varepsilon_{'}}_{h,l}Cl_l \quad \forall h \in \{1,...,k\}
\end{equation}
After the alignment of each class $C_h$ with its corresponding cluster $Cl_l$ has been done as suggested in (23), a classifier $\eta$ can be learned on the transported labeled data $\widetilde{S}= \cup_{q=1}^k \, \widetilde{C_q}$ and evaluated on the unlabeled target data $T$. The proposed HOT-DA approach is formally summarized in Algorithm $1$:
\begin{algorithm}[h]
\caption{HOT-DA}
\label{HOT-DA}
\textbf{Input \qquad}: {$S=\{(x_i,y_i)\}_{i=1}^n, T=\{x_j\}_{j=1}^m$}\\
\textbf{Parameter}: {$\varepsilon, \varepsilon_{'}$}
\begin{algorithmic}[1] 
\STATE Form $\mu_h,\nu_l \quad \forall h,l \in \{1,...,k\}$ (15,16)
\STATE Form $\phi,\varphi$  (17)
\STATE Solve the HOT-DA problem between $\phi$ and $\varphi$  (18)
\STATE Get the one-to-one matching between structures (20) 
\STATE Transport the source structures to the target ones (23) 
\STATE Train a classifier $\eta$ on $\widetilde{S}$ and evaluate it on $T$
\STATE \textbf{return} {$\{y_j\}_{j=1}^m$}
\end{algorithmic}
\end{algorithm}
\section{Experimental Results}
In this section, we evaluate our method on a toy dataset and two challenging real world visual adaptation tasks\footnote{We make our code and the used datasets publicly available at: \\ \url{https://github.com/HOT-DA/HOT-DA}}.
\subsection{Inter-twinning moons dataset}
In the first experiment, we carry on moons dataset, the source domain is the classical binary two inter-twinning moons centered at the origin (0,0) and composed of 300 instances, where each class is associated to one moon of 150 samples. We consider 7 different target domains by rotating anticlockwise the source domain around its center according to 7 angles. Naturally, the greater is the angle, the harder is the adaptation. The experiments were run by setting $\varepsilon = \varepsilon_{'} = 0.1$, and an SVM with a Gaussian kernel as classifier, whose width parameter was chosen as $\sigma = \frac{1}{2\mathbb{V}}$, where $\mathbb{V}$ is the variance of the transported source samples. Our algorithm is compared to an SVM classifier with a Gaussian kernel trained on the source domain (without adaptation) and three optimal transport based domain adaptation methods, OT-GL \cite{courty2016optimal}, JDOT \cite{NIPS2017_0070d23b} and MADAOT \cite{dhouib2020margin}, with the hyperparameter ranges suggested by their authors. To assess the generalization ability of the compared methods, they are tested on an independent set of 1000 instances that follow the same distribution as the target domain. The experiments are conducted 10 times, and the average accuracy is considered as a comparison criterion. The results are presented in Table 1.
\begin{table}[H]
\centering
\setlength\extrarowheight{0.pt}
\setlength\tabcolsep{2.5pt}
\begin{tabular}{lccccccc}
\hline
Angle ($^{\circ}$)  \quad   & 10$^{\circ}$ \quad & 20$^{\circ}$\quad & 30$^{\circ}$\quad & 40$^{\circ}$ \quad & 50$^{\circ}$ \quad & 70$^{\circ}$ \quad & 90$^{\circ}$ \\  
\toprule 
SVM        & \textbf{1}       & 0.896   & 0.760   & 0.688  & 0.600   & 0.266   & 0.172\\ 
OT-GL      & \textbf{1}       & \textbf{1}       & \textbf{1}       & 0.987   & 0.804   & 0.622   & 0.492\\ 
JDOT       & 0.989   & 0.955   & 0.906   & 0.865   & 0.815   & 0.705   & 0.600\\ 
MADAOT     & 0.995   & 0.993   & 0.996   & 0.996   & 0.989   & 0.770   & 0.641\\
\toprule
\textbf{HOT-DA}     & \textbf{1}       & \textbf{1}       & \textbf{1}       & \textbf{1}       & \textbf{1}       & \textbf{0.999}   & \textbf{0.970}\\ 
\toprule
\end{tabular}
\caption{Average accuracy over 10 realizations in moons dataset for 7 rotation angles}
\end{table}
\noindent We remark that all the considered algorithms based on optimal transport manage to achieve an almost perfect score on the angles from $10°^{\circ}$ to $40^{\circ}$, which is rational, as for these small angles the adaptation problem remains quite easy. However, the SVM without adaptation has experienced a decline of almost one third of its accuracy from $30°^{\circ}$. Which proves that moons dataset presents a difficult adaptation problem that goes beyond the generalization ability of standard supervised learning models. For the strongest deformation, from $50^{\circ}$ and up to $90^{\circ}$, our method always provides an almost perfect score, while a considerable deterioration in the performance of OT-GL and JDOT from $50^{\circ}$ and also of MADAOT from $70^{\circ}$ was observed. In short, structures leveraged by HOT-DA are highlighted by eliminating the increasing difficulty of this adaptation task, the constancy of the excellent performances of our approach speaks for itself.
\begin{figure}[H]
	\centering
\includegraphics[width=0.48\textwidth]{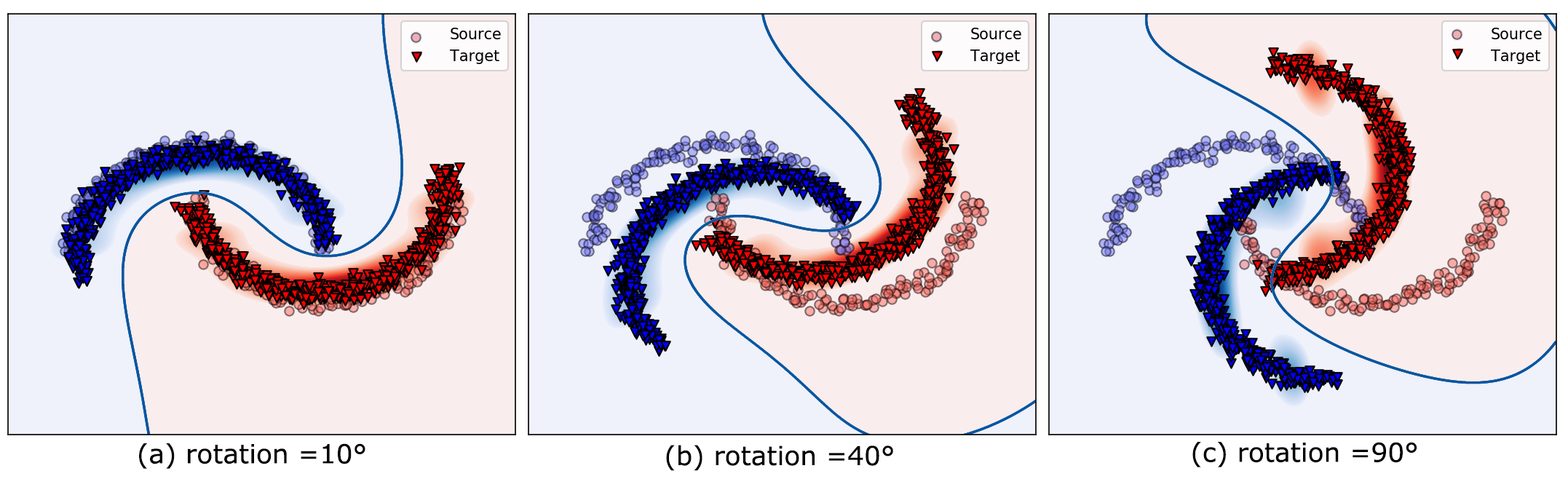}
	\caption{Illustration of the decision boundary of HOT-DA over moons problem for increasing rotation angles.}
\end{figure}
\subsection{Visual adaptation datasets}
We now evaluate our method on two challenging problems. We start by presenting the datasets, the experimental setting, and finish by providing and discussing the obtained results. 
\\
\\ \textbf{Datasets:} We consider two visual adaptation problems. A detailed description of each problem is given in Table 2. 
\begin{table}[H]
\small
\centering
\setlength\extrarowheight{0.pt}
\setlength\tabcolsep{1.5pt}
\begin{tabular}{lcccccc}
\hline
Problem  \quad   & Domains \quad & Dataset \quad & \#Samples \quad & \#Features \quad & \#Classes \quad & Abbr. \\  
\toprule 
Digits & \makecell{USPS \\ MNIST}   & \makecell{USPS \\ MNIST}    & \makecell{1800 \\ 2000}    & \makecell{256 \\ 256}    & \makecell{10 \\ 10}   &\makecell{U \\ M}   \\ 
\toprule 
Objects & \makecell{Caltech\\Amazon\\Webcam\\DSLR }  & \makecell{Caltech\\Office\\Office\\Office }  & \makecell{1123\\958\\295\\157}    & \makecell{4096\\4096\\4096\\4096}   & \makecell{10\\10\\10\\10}  &  \makecell{C\\A\\W\\D} \\ 
\toprule
\end{tabular}
\caption{Description of the visual adaptation problems}
\end{table}
\noindent \textbf{Experimental protocol and hyper-parameter tuning:} For the first problem of digits recognition, 2000 and 1800 images are randomly selected respectively from the original MNIST and USPS datasets. Then, the selected MNIST images  are resized to the same 16 × 16 resolution as USPS ones. For the second problem of objects recognition, Caltech-Office dataset is used, where we randomly sampled a collection of 20 instances per class from each domain, except for DSLR where only 8 instances per class are selected. For this problem, 4096 DeCaf6 features are used to represent the images \cite{donahue2014decaf}. As a classifier for our approach we use 1-Nearest Neighbor classifier (1NN). The comparison is then conducted using 1NN classifier (without adaptation) and four domain adaptation methods, SA \cite{fernando2013unsupervised} with a linear SVM, JDA \cite{long2013transfer} with 1NN classifier, OT-GL with 1NN classifier \cite{courty2016optimal} and JDOT with a linear SVM \cite{NIPS2017_0070d23b}. For digits recognition problem, the experiment was run by setting $\varepsilon = \varepsilon_{'} = 0.5$. As for objects recognition problem, each target domain is equitably splited on a validation and test sets. The validation set is used to select the best hyper-parameters $\varepsilon,\varepsilon_{'}$ in the range of $\{1, ..., 100\}$. The accuracy, is then evaluated on the test set, with the chosen hyper-parameter values. The experimentation is performed 10 times, and the mean accuracy in \% is reported. 
\begin{table}[h]
\small
\centering
\setlength\extrarowheight{0.pt}
\setlength\tabcolsep{2.5pt}
\begin{tabular}{lcccccc}
\hline
Task \quad   & 1NN  \quad & JDA \quad & SA \quad & OT-GL \quad & JDOT  \quad & \textbf{HOT-DA} \\  
\toprule 
M $\to$ U \quad  & 58.33 \quad & 60.09 \quad & 67.71  \quad & 69.96  \quad & 64.00  \quad & \textbf{74.94} \\ 
U $\to$ M \quad  & 39.00 \quad & 54.52 \quad & 49.85  \quad & 57.85  \quad & 56.00  \quad & \textbf{63.30} \\ 
\hdashline
average     \quad  & 48.66 \quad & 57.30 \quad & 58.73  \quad & 63.90   \quad & 60.00  \quad & \textbf{69.12} \\ 
\toprule
A $\to$ C \quad  & 22.25 \quad & 81.28 \quad & 79.20   \quad & \textbf{85.51}   \quad & 85.22   \quad & 83.33 \\ 
A $\to$ D \quad  & 20.38 \quad & 86.25 \quad & 83.80   \quad & 85.00   \quad & 87.90   \quad & \textbf{93.33} \\ 
A $\to$ W \quad  & 23.51 \quad & 88.33 \quad & 74.60   \quad & 83.05   \quad & 84.75   \quad & \textbf{95.10} \\ 
C $\to$ A \quad  & 20.54 \quad & 88.04 \quad & 89.30   \quad & 92.08   \quad & 91.54   \quad & \textbf{92.14} \\ 
C $\to$ D \quad  & 19.62 \quad & 84.12 \quad & 74.40   \quad & 87.25   \quad & 89.91   \quad & \textbf{90.65} \\ 
C $\to$ W \quad  & 18.94 \quad & 79.60 \quad & 88.50   \quad & 84.17   \quad & 88.81   \quad & \textbf{94.69} \\ 
D $\to$ A \quad  & 27.10 \quad & 91.32 \quad & 79.00   \quad & \textbf{92.31}   \quad & 88.10   \quad & 91.33 \\ 
D $\to$ C \quad  & 23.97 \quad & 81.13 \quad & \textbf{92.25}   \quad & 84.11   \quad & 84.33   \quad & 78.28 \\ 
D $\to$ W \quad  & 51.26 \quad & 97.48 \quad & 79.20   \quad & 96.29   \quad & 96.61   \quad & \textbf{97.50} \\ 
W $\to$ A \quad  & 23.19 \quad & 90.19 \quad & 55.00   \quad & 90.62   \quad & 90.71   \quad & \textbf{90.92} \\ 
W $\to$ C \quad  & 19.29 \quad & 81.97 \quad & \textbf{99.60 }  \quad & 81.45   \quad & 82.64   \quad & 72.04 \\ 
W $\to$ D \quad  & 53.62 \quad & \textbf{98.88} \quad & 81.65   \quad & 96.25   \quad & 98.09   \quad & 95.32 \\ 
\hdashline
average      \quad  & 28.47 \quad & 86.72 \quad & 81.65   \quad & 88.18   \quad & 89.05   \quad & \textbf{89.54}\\ 
\toprule
\end{tabular}
\caption{Accuracy over the two visual adaptation datasets}
\end{table}
\\
\\ \textbf{Results:} The results of our experiments are reported in Table 3. From this table, we see that HOT-DA outperforms the other methods on 9 out of 14 tasks, and has the second best accuracy on another task. The  table also  present the  average  results  of  each  algorithm, which confirm the superiority of our approach over the other methods in the two problems. Therefore, we attribute this performance gain to the effectiveness of our Wasserstein-Spectral clustering that succeeds in learning hidden structures in the target domain even if they do not have compact and globular shapes, which is the case in these two challenging visual adaptation problems. Furthermore, the hierarchical formulation efficiently incorporates these structures, which allows to preserve compact classes during the transportation and limits the mass splitting across different target structures.
\section{Conclusion}
In this paper we proposed HOT-DA, a novel approach dealing with unsupervised domain adaptation, by leveraging the ability of hierarchical optimal transport to induce structural information directly into the transportation process. We also proved the equivalence between spectral clustering and the problem of learning probability measures through Wasserstein barycenter, this latter was used to learn hidden structures in the unlabeled target domain as a seminal step before performing hierarchical optimal transport. The algorithm derived from the established approach, has proved to be efficient on both simulated and real-world problems compared to several state-of-the-art methods. In the future, we plan to improve the efficiency of HOT-DA by a deep learning extension to handle larger and more complex datasets.
\small
\bibliography{aaai22.bib}
\end{document}